# A Deep Neural Network Approach for Crop Selection and Yield Prediction in Bangladesh

*Tanhim Islam, Tanjir Alam Chisty, Amitabha Chakrabarty*
BRAC University, Mohakhali, 66 Bir Uttam A.K Khandakar Road, Dhaka- 1212, Bangladesh.
{ tanhimislam, tchisty }@gmail.com, amitabha@bracu.ac.bd

*Abstract*— **Agriculture is the essential ingredients to mankind which is a major source of livelihood. Agriculture work in Bangladesh is mostly done in old ways which directly affects our economy. In addition, institutions of agriculture are working with manual data which cannot provide a proper solution for crop selection and yield prediction. This paper shows the best way of crop selection and yield prediction in minimum cost and effort. Artificial Neural Network is considered robust tools for modeling and prediction. This algorithm aims to get better output and prediction, as well as, support vector machine, Logistic Regression, and random forest algorithm is also considered in this study for comparing the accuracy and error rate. Moreover, all of these algorithms used here are just to see how well they performed for a dataset which is over 0.3 million. We have collected 46 parameters such as – maximum and minimum temperature, average rainfall, humidity, climate, weather, and types of land, types of chemical fertilizer, types of soil, soil structure, soil composition, soil moisture, soil consistency, soil reaction and soil texture for applying into this prediction process. In this paper, we have suggested using the deep neural network for agricultural crop selection and yield prediction.**

I. INTRODUCTION

Bangladesh is predominantly an agrarian country. Due to its very fertile land and favorable weather, varieties of crop grow abundantly in this country. Agriculture sector contributes about 17 percent to the country`s Gross Domestic Product (GDP) and employs more than 45 percent of the total labor force [8]. We live in a country where we have more than 70 percent of agricultural land [5]. Farmers are working so hard to add value in our economic growth for which they need to struggle from early morning with a lot of stress on their head just to make our country a better place of living. However, the process of cultivation or irrigation is doing right now by the farmers is still lagging behind and have not met up today's epoch.

Therefore, we have come up with an idea to change the trend and help the farmers to select the crop efficiently and maximize crop yield with minimal cost. In this paper, we are implementing machine-learning algorithms for agricultural zone 28, consisting of Dhaka, Gazipur, Narayanganj, Tangail, Kishoregonj, Mymensingh and Narsingdi where we are going to analyze the most effective crop selection and yield prediction for every specific region of agriculture. We are considering 6 essential major crops of Bangladesh which are Aus rice, Aman rice, Boro rice, wheat, Potato, and Jute. For the maximum agriculture prediction, we are considering about 46 major features including around 0.3 million data from 2008 to 2017, for our machine learning model to give our farmers the maximum cost-effective farming solution. We have used the artificial neural network to develop diverse crop yield prediction [1]. Under this neural network, we have used the deep neural network with multiple hidden layers which has given the best accuracy result. Furthermore, we have used Support vector machine algorithm, logistic regression and Random forest algorithm to compare with other algorithms.

Meteorological parameters such as average rainfall, maximum temperature, minimum temperature, humidity are exceedingly impacted to our dataset [3]. We have also considered four important chemical fertilizers which are Urea, Triple superphosphate, Diammonium phosphate, and MP. Moreover, we have categorized the land type (Inundation-land highland, Inundation-land medium highland, Inundation-land medium low land, Inundation-land low land, Inundation-land very low land) based on their elevation above the sea level. Furthermore, soil structure is also an important factor, which refers to the shape of soil structural unit. Based on this structure we may get to know in which land what types of crops yield most. The soil type is a very dominant factor for cultivating and efficiently producing a crop [2]. We have considered 19 essential soil type (Calcareous Alluvium, Noncalcareous Alluvium, Acid Basin Clay, Calcareous brown floodplain soil, Calcareous grey floodplain soil, Calcareous dark grey floodplain soil, Noncalcareous grey floodplain soil, Noncalcareous dark grey floodplain soil, Peat, Made-land, Noncalcareous brown floodplain soil, Shallow red-brown terrace soil, deep red-brown terrace soil, Brown mottled terrace soil, Shallow grey terrace soil, deep grey terrace soil, grey valley soil, Brown hill soil, grey valley soil) for our paper. The ideal soil should be considered to be a loam, which means, that particular soil has an equal proportion of sand, silt, and clay. Therefore, based on these we have considered the soil moisture, texture, consistency and its reaction for better visibility of individual soil types.

We have inputted over 0.3 million datasets under 46 factors which are the noble contribution in Agricultural research. To obtain less percentage error, we have done raw data processing, data cleaning and data normalization which drives to the maximum efficiency of our prediction and thus it will help our farmers to boost up the crop yield with minimum effort. As well as, we have applied the deep neural network approach which eventually introduces a new performance benchmark of our entire system. We have perceived a highly promising result with an error percentage of less than 10%. This amount of high accuracy will finally benefit communities to make a better agricultural decision in future days. This will be eventually important for the farmers, government, agricultural stakeholders, policymaker and the society at large in our country towards monitoring food security and for outlining crop yield and business.

II. BACKGROUND STUDY

After a thorough background work, some of the most valuable recent documents and papers are: B. JI et al [11] developed agricultural management need accurate and simple estimation techniques to predict rice yields in the planning process. The intention of the present study where to: (1) observe whether artificial neural network ( ANN) models could effectively predict Fujian rice yield for typical climate condition of the mountain region, (2) evaluate ANN model performance relative to variations of development parameters and (3) compare the effectiveness of multiple linear regression models with ANN models. In this thesis we have described the development of deep neural network model as an alternate and more accurate technique for yielding prediction.

T. Sidique et al [12] introduce a system that is focused on the climate and geographical condition of different areas of Bangladesh. It predicts cost effective crop using a prediction based algorithm. The algorithm are aimed to use is multiple linear regression with the association of some independent variable i.e. rainfall, average maximum temperature and average minimum temperature of certain location and give prediction based on yield rate per unit area. Later, KNNR algorithm was used to compare the accuracy and error rate of the prediction yield rate.

M.S S. Dahika et al [13] uses crop prediction methodology is used to predict the suitable crop by sensing various parameter of soil and also parameter related to atmosphere. For that purpose they used artificial neural network (ANN). This project shows the ability of artificial neural network technology to be used for the approximation and prediction of crop yields at rural district.

D.L. Ehret et al [14] introduce all crop attributes responded in much the same way to individual climate factors. Radiation and temperature generally induced strong positive responses while RH produces negative responses. In the Neural Network models, radiation and temperature were still prominent, but the importance of $CO_2$ in predicting a crop response increased. One advantage of these automated systems is that they offer continuous information across a range of timescales. Furthermore, these systems can readily be used in commercial setting where they can subsequently be improved over time.

B. Lobell et al [15] examines the ability of statistical models to predict yield responses to changes in mean temperature and precipitation, as simulated by a process-based crop model. The CERES-Maize model was first used to simulate historical maize yield variability at nearly 200 sites in Sub-Saharan Africa, as well as the impacts of hypothetical future scenarios of 2 °C warming and 20% precipitation reduction. Statistical models of three types (time series, panel, and cross-sectional models) were then trained on the simulated historical variability and used to predict the responses to the future climate changes. The agreement between the process-based and statistical models' predictions was then assessed as a measure of how well statistical models can capture crop responses to warming or precipitation changes. The performance of statistical models differed by climate variable and spatial scale, with time-series statistical models ably reproducing site-specific yield response to precipitation change, but performing less well for temperature responses. In contrast, statistical models that relied on information from multiple sites, namely panel and cross-sectional models, were better at predicting responses to temperature change than precipitation change.

## III. MATERIAL AND METHOD

### A. Six Major Crops

According to Bangladesh Bureau of Statistics six major crops in Bangladesh are Aus rice, Aman rice, Boro rice, Jute, Wheat, and Potato.

### B. Sample Area

We are considering agricultural zone-28 for our prediction model which includes:
- Dhaka
- Gazipur
- Mymensingh
- Narayangonj
- Tangail
- Kishoregonj
- Narshingdi

### C. Dataset

Regardless of the approach taken, data collection and proceedings are always being diligent and time-consuming. Online surveys can be used to get to know about the circumstances of the area under study but it would be incomplete if we are unable to search in person. Depending on the design of our data collecting survey, we have not only gone for web resources yet we tried out to communicate with some established organizations which are precisely working to boost up the agricultural level. Before starting the research work on this sector our main barrier was to collect data. That is why we at first tried to do some online surveys. But there was lack of enough resources that we actually needed. So we have moved on to the next step which was to collect data from the different institute. We first moved to Soil Research Development Institute (SRDI), which works with soil nutrients. We were working on agricultural zone 28 so we had collected about 70 books. The first problem was all the data were in manual form, which we needed to make digitalized. This took a long process to manually insert the data in the computer. Next on we had to move on to the Bangladesh Agricultural Research Council (BARC). We again take the bookish data from that institute and inserted the data manually into the computer. From SRDI and BARC we collected about 6 different types of land Inundation-land highland, Inundation-land medium highland, Inundation-land medium low land, Inundation-land low land, Inundation-land very low land, miscellaneous land. We also have collected 4 different types of Chemical fertilizer. These are Urea, Triple Super Phosphate, Diammonium Phosphate, and MP. We also have considered 19 different types of soil and 4 different types of soil information. For web resources, we took the initiative to go with Bangladesh Bureau of Statistics Government (BBS) and Bangladesh Meteorological Department (BMD) data. From BMD we have collected average rainfall, maximum temperature, minimum temperature, and humidity data from (2008-2017). From BBS we have collected the agricultural output data which consists of crop yield, area of individual crops and production of individual crops. As agricultural zone number 28 consists of Dhaka, Narayanganj, Gazipur, and Tongi so it would become easier for us to visit those areas in person and investigate it so we decided to work on this zone.

We collected about 0.3 million combined data along with 46 important features from 2008 to 2017.

*D. Deep Neural Network*

In our thesis, we have implemented Back propagation method to train our deep neural network where we have used 3 hidden layers to calculate the total cost of our output. Sometimes handling enormous amount data by using only 1 hidden layer may not give a high accuracy. Although adding more hidden layers may increase computational cost but it can better generalize to big data.

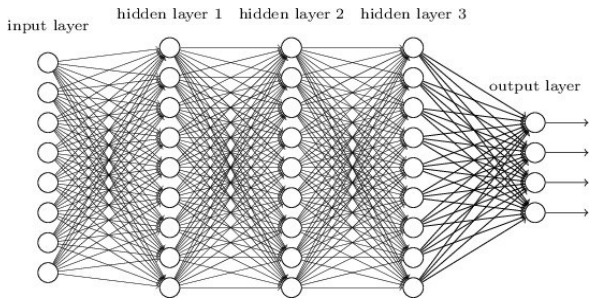

Figure 1. A simple Deep Neural Network model

Pseudo code for Back propagation to train Deep Neural Network

**Algorithm 1:** Back Propagation Deep Neural Network

1. Initialize network weights (often small random values)
2. **do**
3.     **for Each** training example ex
4.         Prediction = neural-net-output(network, ex) // *forward pass*
5.         Actual =teacher-output(ex)
6.         Compute error (prediction - actual) at the output units
7.         Compute $\Delta w_h$ for all weights from hidden layer to output layer // *back pass*
8.         Compute $\Delta w_i$ for all weights from input layer to hidden layer // *back pass*
9.         Update network weights // *input layer not modified by error estimate*
10. **until** all examples classified correctly or another stooping criterion satisfied
11. **return** the network

IV. DATASET AND RESULT ANALYSIS

At first, we clean the unwanted observation from the dataset what we have acquired so far. There might have been duplicate data because of the maximum amount of data was on hard copy so it is very obvious that one data might be inserted twice. There are also some irrelevant observations which actually don't fit with the specific problem that we are trying to solve.

TABLE I    Dataset sample of few features out of 46

| District | year | avgrainfall | humidity | urea | tsp | DAP |
|---|---|---|---|---|---|---|
| dhaka | 2008 | 2385 | 71 | 25967 | 8262 | 1573 |
| dhaka | 2009 | 1930 | 66 | 29300 | 8041 | 1578 |
| dhaka | 2010 | 1523 | 55.3 | 27967 | 8675 | 1688 |
| dhaka | 2011 | 1776 | 68.2 | 36991 | 9386 | 1628 |
| dhaka | 2012 | 1329 | 58 | 32966 | 8166 | 1925 |
| dhaka | 2013 | 1590 | 63 | 30332 | 9188 | 1941 |
| dhaka | 2014 | 1399 | 64.7 | 35321 | 9233 | 1926 |
| dhaka | 2015 | 2166 | 60.2 | 30750 | 9544 | 1586 |
| dhaka | 2016 | 1562 | 65 | 36712 | 9094 | 1987 |
| dhaka | 2017 | 1640 | 69 | 37382 | 8782 | 1903 |
| gazipur | 2008 | 2197 | 71 | 29264 | 9225 | 5721 |
| gazipur | 2009 | 1912 | 66 | 29059 | 9658 | 5604 |
| gazipur | 2010 | 1181 | 55.3 | 29295 | 9435 | 5967 |
| gazipur | 2011 | 1777 | 68.2 | 29457 | 9866 | 5831 |

*A. Dataset Analysis*

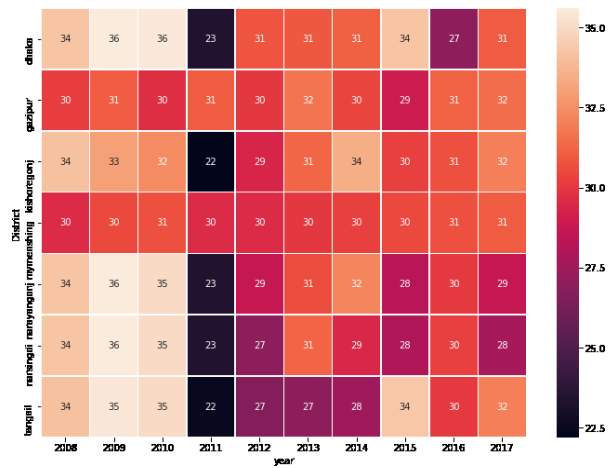

Figure 2. Maximum Temperature

In the figure 2. We have described the utmost temperature from year 2008 to 2017 in Dhaka, Gazipur, Narayangonj, Tangail, Kishorgonj, Mymenshing, and Narsingdi district.. In the horizontal axis, we have set year (2008-2017) and in the vertical axis we have set district. Annual maximum temperature was recorded from 22.5 degree Celsius to 35 degree Celsius. As we can see in Dhaka, city during 2008 it was about 34 degree Celsius and almost most of the 7 districts are in between 30 to 34 degree Celsius. But as we move forward the temperature increases 1 to 2 units in almost every district till 2010. But the temperature drastically got down during 2011 expects Gazipur and Mymenshing. Eventually the temperature again increases and being stable from year 2012 to 2017.

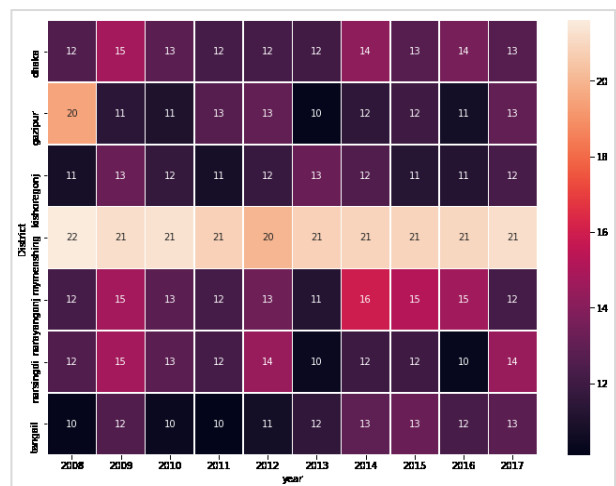

Figure 3. Minimum Temperature

Annual minimum temperature was recorded from 12 degree Celsius to 20 degree Celsius. As we can see that in Mymenshing district minimum temperature is recorded between 20 degree Celsius to 22 degree Celsius which is higher that other districts.Also, in 2009, annual minimum temperature increases 3 unit in Dhaka, narayangonj and narsingdi district.Furthermore, in Dhaka district minimum temerature ranges from 12 degree Celsius to 15 degree Celsius, in Gazipur district minimum temperature ranges from 11 degree Celsius to 20 degree Celsius, in Kishoregonj district minimum temperature ranges from 11 degree Celsius to 13 degree Celsius, in Mymenshing district minimum temperature ranges from 20 degree Celsius to 22 degree Celsius, in Narayangonj district minimum temperature ranges from 10 degree Celsius to 15 degree Celsius and in Narsingdi district it varies from 10 degree Celsius to 13 degree Celsius.

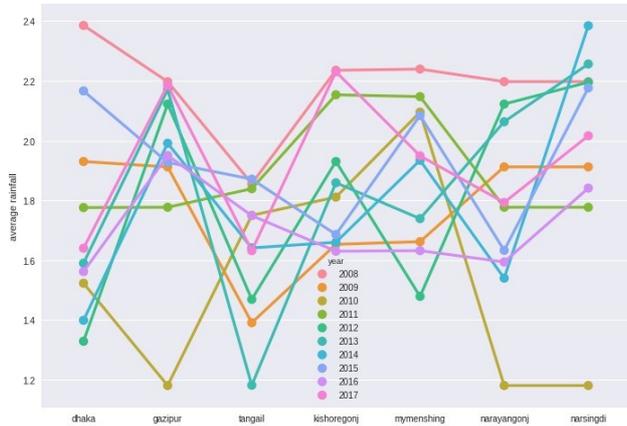

Figure 4. Average Rainfall

Annual average rainfall was recorded from 12 mm to 24 mm. As we can visualize that Dhaka and Narsingdi have received the highest amount of rainfall during 2008 and 2014, which was about 24mm and performs the least in 2012 and 2014 which was about in between 12mm and 14mm. The most static rainfall happened in Kishoregonj district where the average rainfall was about 16mm to 22.3 mm. But in 2010 almost all of the districts got the least amount of rainfall. This resulted because during that time the annual average maximum temperature was really high in the entire district so the precipitation was really poor.

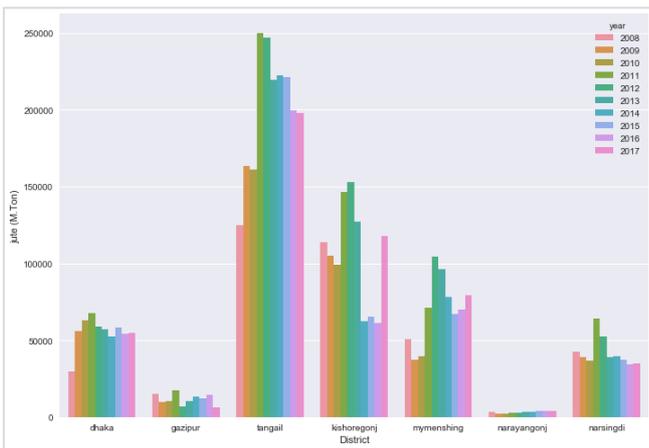

Figure 5. Production of Jute rice

In figure 5. We have represented the result analysis of the production of Jute from the year 2008 to 2017 in Dhaka, Gazipur, Narayanganj, Tangail, Kishoregonj, Mymenshing, and Narsingdi district [10]. In the horizontal axis, we have set districts (Agricultural Zone-28) and in the vertical axis, we have set Jute rice production (metric ton). As we can see from the above figure that Tangail is the leading district with highest production whereas Mymenshing has low amount of yielding rate. Types of land found in Tangail are consists of different pattern of soil which are, non-calcareous alluvium, acid basin clay, calcareous grey floodplain soil, non-calcareous grey floodplain,non-calcareous grey floodplain, deep red-brown soil, deep grey terrace soil, grey valley soil. Reasoning for this highest rate of production in Tangail, is Jute is a thirsty plant and Tangail has the significant rainfall with average high temperature and humidity. In addition to these soil of Tangail has best soil moisture among other three types soil information. Moreover the fertilizers were high in Mymenshing and less soil moisture than soil texture which is good for rice production but not for fertility-exhausting plant like jute.

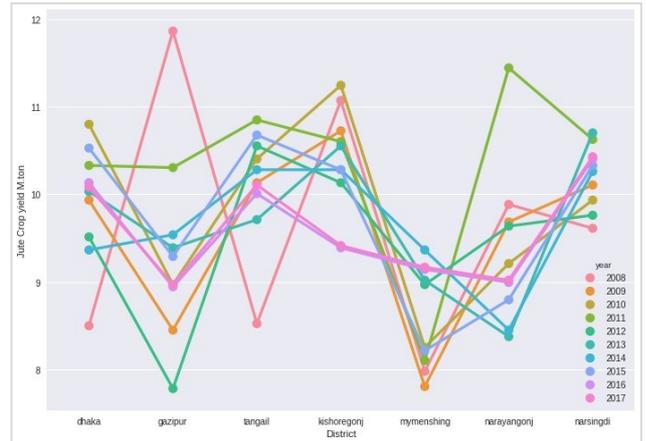

Figure 6: Crop yield analysis of jute

In figure 6. We have described the crop yield of Jute from the year 2008 to 2017 in Dhaka, Gazipur, Narayangonj, Tangail, Kishorgonj, Mymenshing, and Narsingdi district. This is the agricultural output of Jute. In the horizontal axis, we have set districts (Agricultural Zone-28) and in the vertical axis, we have set crop yield hector (metric ton) for Jute. From the figure 5.8.1 Tangail has highest production but slightly decreased after 2015. For this in figure 5.8.2 Kishoregonj and Narsingdi has a good amount of yielding rate because of stable required environment, atmosphere, soil and fertilizer. Furthermore from the graph we can see that ratio of Gazipur rate over the year is more fluctuating. 2008 Gazipur has the highest amount of yielding rate (11.8M.ton) but it decreased drastically in 2009. In 2009 temperature was very low near to 11 degree Celsius which is not feasible for keeping the rate same as 2008.

### A. Performance Analysis

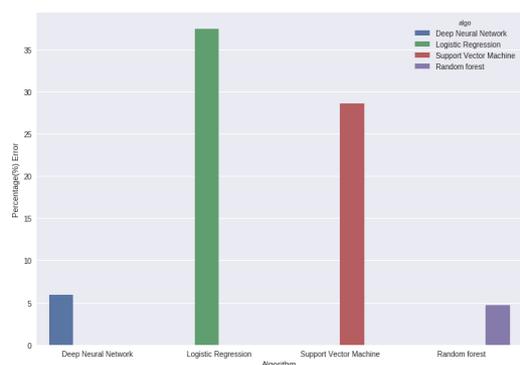

Figure 7: Percentage Error of Jute for different algorithms

In figure 7, it is showing that deep neural network is performing much better than the logistic regression and support vector machine. Random forest thus has error under 10 percent. In the horizontal axis, we have set algorithm and in the vertical axis, we have set percentage (%) error. For Jute dataset, we have considered 80% training set and 20% as testing set, for all four algorithms. In addition, we got 94.1% accuracy for deep neural network with 5.9 % mean square error, 71.4% accuracy for support vector machine with 28.63% mean square error, 95.3% accuracy for random forest with 4.7% mean square error and 62.56% accuracy for logistic regression with 37.44% mean square error. Thus, we are choosing deep neural network approach for crop selection and yield prediction.

TABLE II   Evaluation measures of Aus rice

| Method | Training (%) | Testing (%) | Accuracy (%) | MSE (%) |
|---|---|---|---|---|
| Deep Neural Network(DNN) | 80% | 20% | 97.7% | 2.3% |
| Support Vector Machine(SVM) | 80% | 20% | 73.3% | 26.7% |
| Random Forest | 80% | 20% | 90.7% | 9.3% |
| Logistic Regression | 80% | 20% | 52.57% | 47.43% |

TABLE III   Evaluation measures of Aman ric

| Method | Training (%) | Testing (%) | Accuracy (%) | MSE (%) |
|---|---|---|---|---|
| Deep Neural Network(DNN) | 80% | 20% | 94.6% | 5.4% |
| Support Vector Machine(SVM) | 80% | 20% | 69.7% | 30.3% |
| Random Forest | 80% | 20% | 92.1% | 7.9% |
| Logistic Regression | 80% | 20% | 49.3% | 50.7% |

TABLE IV   Evaluation measures of Boro rice

| Method | Training (%) | Testing (%) | Accuracy (%) | MSE (%) |
|---|---|---|---|---|
| Deep Neural Network(DNN) | 80% | 20% | 96.7% | 3.3% |
| Support Vector Machine(SVM) | 80% | 20% | 67% | 33% |
| Random Forest | 80% | 20% | 91.2% | 8.8% |
| Logistic Regression | 80% | 20% | 56.11% | 43.89% |

TABLE V   Evaluation measures of Potato

| Method | Training (%) | Testing (%) | Accuracy (%) | MSE (%) |
|---|---|---|---|---|
| Deep Neural Network(DNN) | 80% | 20% | 97.3% | 2.7% |
| Support Vector Machine(SVM) | 80% | 20% | 65% | 35% |
| Random Forest | 80% | 20% | 88.3% | 11.7% |
| Logistic Regression | 80% | 20% | 78% | 22% |

TABLE VI   Evaluation measures of Wheat

| Method | Training (%) | Testing (%) | Accuracy (%) | MSE (%) |
|---|---|---|---|---|
| Deep Neural Network(DNN) | 80% | 20% | 96% | 4% |
| Support Vector Machine(SVM) | 80% | 20% | 67% | 33% |
| Random Forest | 80% | 20% | 90% | 10% |
| Logistic Regression | 80% | 20% | 75% | 25% |

TABLE VII   Evaluation measures of Jute

| Method | Training (%) | Testing (%) | Accuracy (%) | MSE (%) |
|---|---|---|---|---|
| Deep Neural Network(DNN) | 80% | 20% | 94.1% | 5.9% |
| Support Vector Machine(SVM) | 80% | 20% | 71.4% | 28.6% |
| Random Forest | 80% | 20% | 95.3% | 4.7% |
| Logistic Regression | 80% | 20% | 62.56% | 37.44% |

V. CONCLUSION

To sum up, in our research we tried to develop a model in the agricultural sector by using the 21st century's modern approach. Our study tried to give a prediction on how a simple machine learning algorithm can change our country's agriculture image. Being dependent on the agricultural side for a long time our country has not meet much between agriculture and technology so far. Although there are some mobile apps, which are being, build but which is not actually up to the mark. Right now, the people of our generation are in a position where everyone is in touch with modern things. So it is high time we should aim at a future to live a better lifestyle. Our government has already taken so many good initiatives in the agricultural sector. It is high time to precede digitally in this sector so that; not only the government but also stockholder and society might get benefitted out from it. Our one little step will be enough to introduce digital agriculture system for best crop selection and yield prediction.

REFERENCES

[1] Shastry, K., Sanjay, H. &Deshmukh, A. (2016). A Parameter Based Customized Artificial Neural Network Model for Crop Yield Prediction. Journal of Artificial Intelligence,9, 23-32 doi: 10.3923/jai.2016.23.32

[2]Patil, D. &Shirdhonkar, M. (2017). Rice Crop Yield Prediction using Data Mining Techniques: An Overview. International Journal of Advanced Research in Computer Science and Software and Software Engineering, 7(5), 427-431

[3] Rahman, M., Haq, N. &Rahman, R. (2014). Machine Learning Facilitated Rice Prediction in Bangladesh. Annual Global Online conference on Information and Computer Technology, 5,1-4.doi: 10.1109/GOCICT.2014.9

[4] Trading Economics. (2018). Bangladesh GDP Growth Rate. Retrieved from https://tradingeconomics.com/bangladesh/gdp-growth

[5] Trading Economics. (2015). Bangladesh-Agricultural land(% of land area).Retrieved from https://tradingeconomics.com/bangladesh/agricultural-land-percent-of-land-area-wb-data.html

[6] Trading Economics. (2016). Bangladesh-Agricultural value added (annual % growth). Retrieved from https://tradingeconomics.com/bangladesh/agriculture-value-added-annual-percent-growth-wb-data.html

[7] Svozil, D., Kvanička, V., &Pospichal, J. (1997). Introduction to multi-layer feed forward neural networks. Chemometrics and intelligent laboratory systems, 39, 43-62.

[8] Bangladesh Bureau of Statistics. (2008-2017). Bangladesh Bureau of Statistics. Retrieved from http://www.bbs.gov.bd/.

[9] Bangladesh Meteorological Department. (2008-2017).Bangladesh Meteorological Department. Retrieved from http://www.bmd.gov.bd/

[10] Soil Resource Development Institute. (2008-2017). Soil Resource Development Institute. Retrieved from http://www.srdi.gov.bd/

[11] B. J I ET AL. (2007). Artificial neural networks for rice yield prediction in mountainous regions. *Journal of Agricultural Science* (2007), 145, 249–261.

[12] T. Siddique. (2016). Automated firming prediction. *Department of Computer Science and Engineering, Brac University.*

[13] M.S S.Dahikar. (2014). Agricultural Crop Yield Prediction Using Artificial Neural Network Approach. *International journal of innovative research in electrical, electronics, instrumentation and control engineering*, vol. 2, issue 1

[14] D.L. Ehret et al. (2011). Neural network modeling of greenhouse tomato yield, growth and water use from automated crop monitoring data. *Computers and Electronics in Agriculture*. 82–89.

[15] B. Lobell et al. (2010). On the use of statistical models to predict crop yield responses to climate change. *Agriculture and forest meterology*. Volume 150. Issue 11